\documentclass[10pt,twocolumn,letterpaper]{article}

\usepackage{cvpr}
\usepackage{times}
\usepackage{graphicx}
\usepackage{amsmath}
\usepackage{amssymb}
\usepackage{helvet}
\usepackage{courier}
\usepackage{mathrsfs}
\usepackage{makecell}
\usepackage{mathtools}
\DeclarePairedDelimiter\floor{\lfloor}{\rfloor}
\usepackage{graphicx}
\usepackage{verbatim}
\usepackage{authblk}
\graphicspath{{images/}}
\usepackage{color}

\usepackage[breaklinks=true,bookmarks=false,colorlinks]{hyperref}

\cvprfinalcopy % *** Uncomment this line for the final submission

\def\cvprPaperID{2566} % *** Enter the CVPR Paper ID here

\begin{document}

%%%%%%%%% TITLE
\title{Optical Flow Guided Feature: A Fast and Robust Motion Representation for Video Action Recognition}
%Spatial-Temporal
% \author[1, 2]{Shuyang Sun}
% \author[1]{Zhanghui Kuang}
% \author[3]{Lu Sheng}
% \author[2]{Wanli Ouyang}
% \author[1]{Wei Zhang}
% \affil[1]{SenseTime Research}
% \affil[2]{School of Electrical and Information Engineering, The University of Sydney}
% \affil[3]{Department of Electronic Engineering, The Chinese University of Hong Kong}
\author{
Shuyang Sun$^{1,2}$, \hspace{1px} Zhanghui Kuang$^2$, \hspace{1px} Lu Sheng$^3$, \hspace{1px} Wanli Ouyang$^1$, \hspace{1px} Wei Zhang$^2$\\
$^1$The University of Sydney \hspace{3px} $ ^2$SenseTime Research \hspace{3px} $ ^3$The Chinese University of Hong Kong\\
{\tt\small \{shuyang.sun wanli.ouyang\}@sydney.edu.au \hspace{2px} \{wayne.zhang kuangzhanghui\}@sensetime.com\hspace{2px} lsheng@ee.cuhk.edu.hk}
}
%\{shuyang.sun wanli.ouyang\}@sydney.edu.au}\hspace{2px} \{wayne.zhang kuangzhanghui\}@sensetime.com\hspace{2px} lsheng@ee.cuhk.edu.hk

\maketitle
\begin{abstract}
%\begin{quote}
Motion representation plays a vital role in human action recognition in videos. In this study, we introduce a novel compact motion representation for video action recognition, named Optical Flow guided Feature (OFF), which enables the network to distill temporal information through a fast and robust approach.
The OFF is derived from the definition of optical flow and is orthogonal to the optical flow. The derivation also provides theoretical support for using the difference between two frames.
By directly calculating pixel-wise spatio-temporal gradients of the deep feature maps, the OFF could be embedded in any existing CNN based video action recognition framework with only a slight additional cost. It enables the CNN to extract spatio-temporal information, especially the temporal information between frames simultaneously. This simple but powerful idea is validated by experimental results. The network with OFF fed only by RGB inputs achieves a competitive accuracy of $93.3\%$ on UCF-101, which is comparable with the result obtained by two streams (RGB and optical flow), but is 15 times faster in speed. Experimental results also show that OFF is complementary to other motion modalities such as optical flow. When the proposed method is plugged into the state-of-the-art video action recognition framework, it has $96.0\%$ and $74.2\%$ accuracy on UCF-101 and HMDB-51 respectively. The code for this project is available at: \href{https://github.com/kevin-ssy/Optical-Flow-Guided-Feature}{https://github.com/kevin-ssy/Optical-Flow-Guided-Feature}

%and reduces the error rate by $33.3\%$ compared with the baseline

%For efficiency, it in-explicitly represents feature flow in its dual space. 

%Motion representation plays a significant role in video-based challenges. Current popular motion representations such as optical flow, iDT, or motion vectors are either computationally expensive or unable to achieve high performance. In this paper, we present a novel kind of modality to capture motion information, namely the \textit{Fast Feature Flow (OFF)}. The OFF is not only fast and end-to-end trainable but also able to achieve comparable performance comparing with optical flow. We also find that apart from the feature that physically describes the velocity, the acceleration describing features also matters in video action recognition tasks. Based on the design philosophy of OFF, our method can outperform the state-of-the-art algorithms either in speed or accuracy on two popular action recognition datasets UCF-101 and HMDB-51. Our fastest model can run on 240 fps on a single GPU but still achieve 92.5\% mAP on UCF 101, while the best model could decrease the error rate from 6\% to 5\% on the same dataset. Besides, our feature design philosophy could be further extended on other video tasks apart from action recognition.
%\end{quote}
\end{abstract}

\vspace{-12px}
\section{Introduction}
Video action recognition has received longstanding attentions in the community of computer vision for decades. It aims at automatically recognizing human action from video sequences. Since CNNs have achieved great successes in image classification and other related tasks \cite{krizhevsky2012imagenet,simonyan2014vgg,Szegedycvpr2015googlenet,he2016resnet,zeng2017crafting,Zhao_2017_CVPR,ouyang2016learning}, lots of CNN based methods have been proposed by considering video action recognition as a classification task \cite{carreira2017i3d, wang2016tsn, ng2016actionflownet, zhang2016motionvector, feichtenhofer2016spatiotemporal, diba2016tle, diba20163dtwostream, wang2015tdd, wang2015verydeep, sun2015human, simonyan2014two}. Compared to the image classification methods, temporal information is the key ingredient of video action recognition.

\begin{figure}[t]
\centering
\includegraphics[scale=0.88]{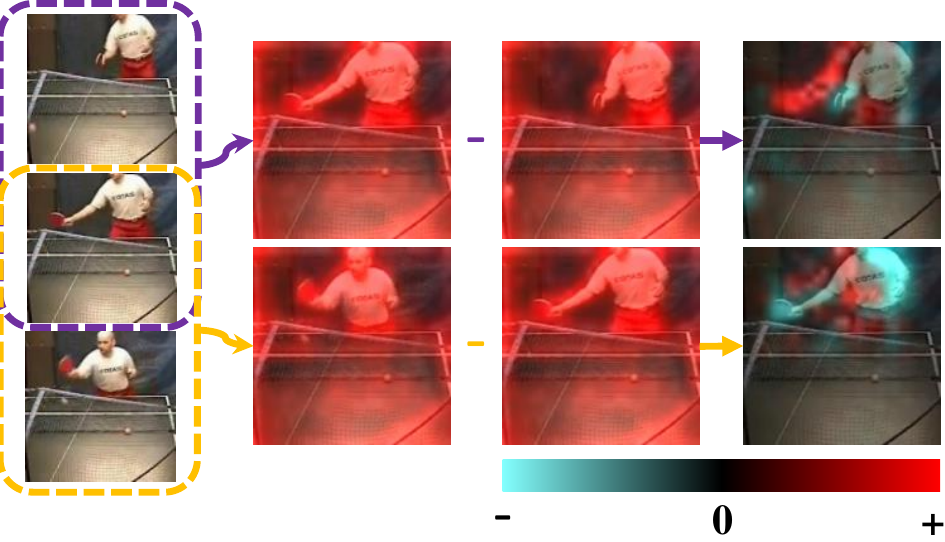}
% \bigskip
\caption{\textbf{The Optical Flow guided Feature (OFF).} Left column: input frames. Middle two columns: standard deep features before applying OFF onto two frames. Right column: temporal difference in OFF. The colors red and cyan are used respectively for positive and negative values. The feature difference between two frames is valid and comprehensive in representing motion information. Best seen in color and zoomed in.}
\label{fig:headpic}
\end{figure}

%Many works \cite{simonyan2014two, wang2016tsn, carreira2017i3d} in action recognition aim at answering the following question \textit{how to design/use motion representation that is both fast and robust?} 
Optical flow is found to be a useful motion representation in video action recognition, including the Two-Stream-based \cite{simonyan2014two, wang2016tsn} and 3D convolution-based methods \cite{carreira2017i3d}.
However, extracting dense optical flows is still inefficient. It costs over 90\% of the whole run-time in a two-stream based pipeline  both at training and testing phases.
Moreover, 3D convolutions on RGB input can also capture temporal information, but the RGB-based 3D CNN still does not perform on par with its two-stream version.
%
% However, extracting the dense optical flow required by the Two-Stream based methods is still inefficient both at training and testing stages (14.75 fps by using a Nvidia Titan X GPU \cite{zach2007tvl1}), which constrains the speed of these methods, and the 3D CNN taking RGB as input still does not perform on par with its two-stream version.
%
Other motion descriptors, e.g., 3DHOG \cite{Klaser2008}, improved Dense Trajectory \cite{wang2013idt}, and motion vector \cite{zhang2016motionvector}, are either inefficient or not so effective as optical flow.
%Other works use 3D CNN and temporal pooling to capture the continuous video-level spatio-temporal information \cite{carreira2017i3d,varol2017ltc,tran2017res3d,diba20163dtwostream}. 

\textit{How to design/use motion representation that is both fast and robust?} To this end, the required computation should be economical and the representation should be sufficiently guided by the motion information. Taking the above requirements into consideration, we propose the Optical Flow guided Feature (OFF), which is fast to compute and can comprehensively represent motion dynamics in a video clip.

In this paper, we define a new feature representation from the orthogonal space of optical flow on the feature level \cite{HornBertholdK.P.;Schunck1981}. Such definition brings the guidance from optical flow here to the representation, therefore, we name it as the Optical Flow guided Feature (OFF). The feature consists of spatial gradients of feature maps in horizontal and vertical directions, and temporal gradients obtained from the difference between feature maps from different frames. Since all the operations in OFF are differentiable, the whole process is end-to-end trainable when OFF is plugged into one CNN architecture. Actually the OFF unit only consists of pixel-wise operators on CNN features. These operators are fast to apply, and enable the network with RGB input to capture spatial and temporal information simultaneously.

One vital component in OFF is the difference between features from different images/segments. As shown in Fig. \ref{fig:headpic}, the difference between the features from two images provides representative motion information that can be conveniently employed by CNNs. The negative values in the difference image depict the locations where the body parts/objects disappear, while the positive values represent where they emerge. This pattern of disappearing at one location and emerging at another location can be easily treated as a specific motion pattern and captured by later CNN layers. The temporal difference could be further combined with the spatial gradients such that the constituted OFF is guided by the optical flow on feature level according to our derivation in later section. Moreover, calculation of the motion dynamics at the feature level is faster and  also more robust because 1) it enables the spatial and temporal networks with the capability of weight sharing and 2) deeply learned features convey more semantic and discriminative representations with reliable elimination of local and background noises in the raw frames.

Our work has two main contributions. 
% two contribution: 1. Fast 2.Robust

First, \textbf{OFF is a fast and robust motion representation}. OFF is fast to enable over 200 frames per second with \textit{only RGB} as the input and is derived from and guided by the optical flow. Taking only RGB from videos, experimental results show that the CNN with OFF is close in performance when compared with the state-of-the-art optical flow based algorithms. The CNNs with OFF can achieve $93.3\%$ on UCF-101 with \textit{only RGB} as the input, which is currently state-of-the-art among the RGB-based action recognition methods. When plugging OFF in the state-of-the-art action recognition method \cite{wang2016tsn} in a Two-Stream manner (RGB + Optical Flow), the performance of our algorithm could result in $96.0\%$ on UCF-101 and $74.2\%$ on HMDB-51.

%Second, \textbf{OFF is robust, applicable on spatial descriptive deep features, and also on many other modalities like optical flow.} To overcome some obstacles like luminance variance or object overlapping, we propose to apply OFF on deep feature level, which is more robust than  OFF is also complementary to some motion descriptors like optical flow, in this case, the OFF may come to be a kind of \textit{acceleration representation}. 

Second, \textbf{an OFF equipped network can be trained in an end-to-end fashion}. In this way, the spatial and motion representations can be jointly learned through a single network. This property is friendly for video tasks on large-scale datasets, as it may not require the network to pre-compute and store motion modalities for training. Besides, the OFF can be used between images/segments  in a video clip both on image level and feature level. %This property allows us to accelerate the testing process as the frames sampled are able to describe a longer range of video. 
%With the capacity to extract temporal information on feature level, the network could save its number of parameters and computational cost by sharing network weights from the spatial and temporal two streams. That is, we somehow combine the independent two streams into one, which also enables the network to capture spatial and temporal representations simultaneously.

%Moreover, \textbf{we provide theoretical support for the use of temporal difference between frames in video based tasks.} Previous works have shown that the temporal difference between frames is useful in video related tasks \cite{wang2016tsn}, however, there is no theoretical evidence to help  explain why this simple idea works that well. In this paper, we will give a theoretical derivation to illustrate the relationship between the temporal difference and optical flow, and suggest a way of using such a modality in video related tasks.

The rest of this paper is organized as follows. Section ~\ref{sec:relate} introduces recent methods that are related to our work. Section~\ref{sec:method} illustrates the definition of OFF and details our proposed method. Section~\ref{sec:cnn} explains our implementation method in CNN. Our experimental results is summarized in section~\ref{sec:experiment}, with concluding remarks in conclusion Section \ref{sec:con}.

\section{Related Work}
\label{sec:relate}

Traditional methods extracted hand-craft local visual features such as 3DHOG \cite{Klaser2008}, Motion Boundary Histograms (MBH) \cite{Dalal2006}, improved Dense Trajectory (iDT) \cite{wang2013idt, wang2011idt} and then encoded them into sparse or compact feature vectors which were fed into classifiers \cite{Scovanner2007,Peng2014}. Deeply learned features were then found to perform better than hand-crafted features for action recognition \cite{simonyan2014two,wang2015tdd}.  

As a significant breakthrough in action recognition, Two-Stream based frameworks used the deep CNN to learn from the hand-craft motion features like optical flow and iDT \cite{simonyan2014two, wang2015tdd, zhang2016motionvector, wang2016tsn, diba20163dtwostream, yue2015beyondshortlstm, carreira2017i3d, tran2015c3d, feichtenhofer2016spatiotemporal, feichtenhofer2017spatio-temp-multiplier}. These attempts have achieved remarkable progress in improving the recognition accuracy, but still rely on the pre-computed optical flow or iDT, which constrains the speed of the whole framework. 
 
In order to obtain the motion modality in a fast way, recent works used optical flow only at the training stage \cite{ng2016actionflownet}, or proposed motion vector as the simplified version of optical flow \cite{zhang2016motionvector}. These attempts have produced degraded optical flow results and still did not perform on par with the approaches using traditional optical flow as the input stream.
%are either not efficient enough or not effective enough, and the CNN-estimated optical flow may also not be complementary to the optical flow itself.

Many approaches learn to capture the motion information directly from input frames using 3D CNN \cite{tran2015c3d, varol2017ltc, carreira2017i3d, tran2017res3d, diba20163dtwostream, varol2017long3dconv}. Boosted by the temporal convolution and pooling operations, 3D CNN could distill the temporal information between consecutive frames without segmenting them into short snippets. Compared with the learning of filters to capture motion information, our OFF is a principled representation mathematically derived from the optical flow. 3D CNN, constrained by network design, training sample, and parameter regularization like weight decay, may not be able to learn good motion representation like OFF. 
Therefore, current state-of-the-art 3D CNN based algorithms still rely on traditional optical flow to help the networks to capture motion patterns. In comparison, our OFF 1) well captures the motion patterns so that RGB stream with OFF performs on par with two stream methods, and 2) is also complementary to other motion representations like optical flow.

%At the same time, the 3D CNN and temporal pooling require more computational cost for a single network, as we need to sample at least 32 frames for a single video in both training and testing in order to maintain the feature pyramid \cite{carreira2017i3d}.

To capture long-term temporal information from videos, one intuitive approach is to introduce the Long Short-Term Memory (LSTM) module as an encoder to encode the relationship between the sequence-illustrating deep features \cite{yue2015beyondshortlstm, sun2017lattice, shi2017shuttlenet}. LSTM can still be applied on the OFF. Therefore, our OFF is complementary to these methods.  %However, the performance of state-of-the-art Two-Stream based LSTM is still somehow lower than the pure CNN based methods when comparing under the same circumstances \cite{sun2017lattice, shi2017shuttlenet}.

Concurrent with our work, another state-of-the-art method applies a strategy called \textit{ranked pool} \cite{fernando2017rankpool} that generates a fast video-level descriptor, namely, the \textit{dynamic images} \cite{bilen2016dynamic}. However, the very nature in design and implementation between the dynamic images and ours are different. The dynamic images are designed to summarize a series of frames while our method is designed to capture the motion information related to optical flow.

\section{Optical Flow Guided Feature}

\begin{figure}[t]
% \centering
\includegraphics[scale=0.5]{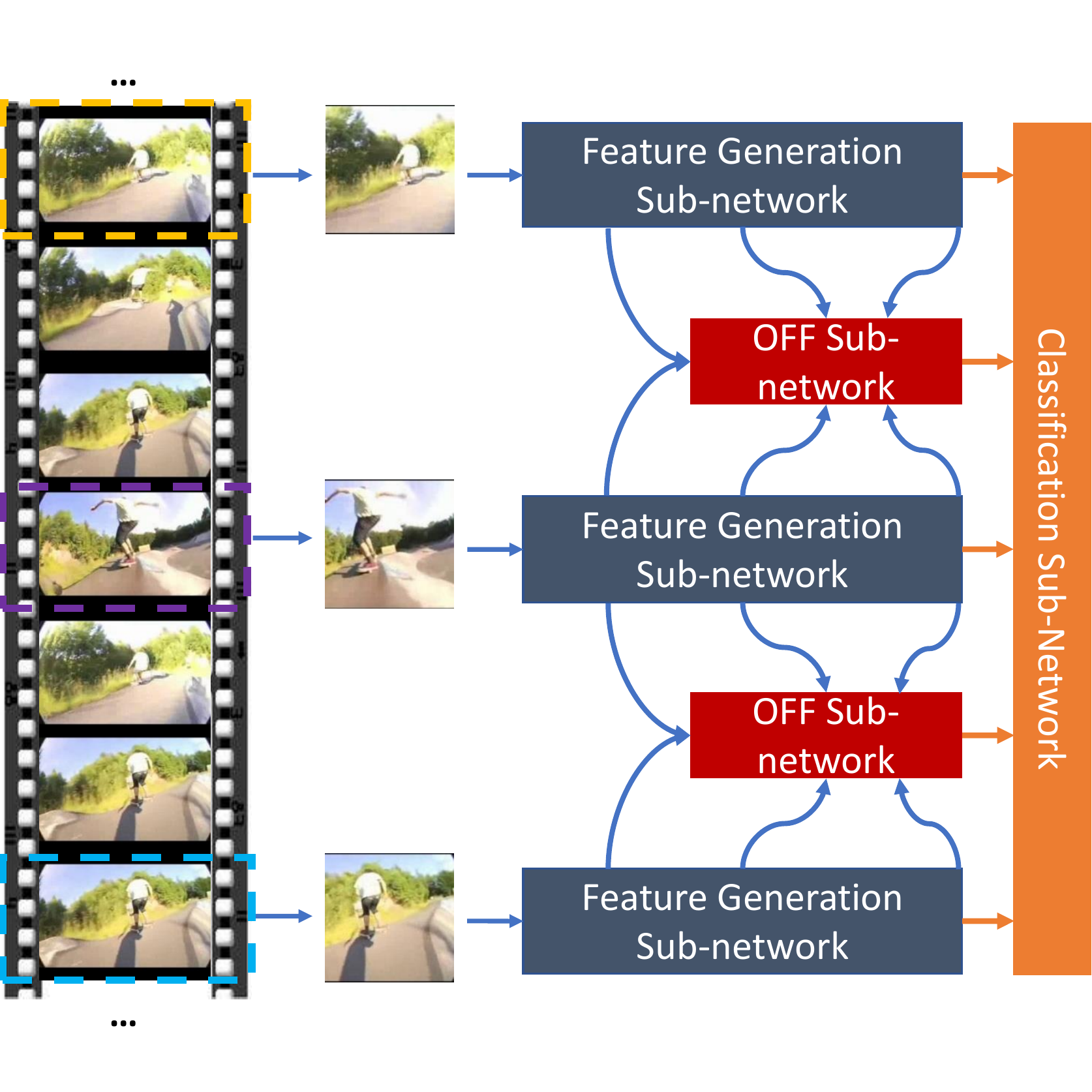}
\caption{\textbf{Network architecture overview.} The feature generation sub-network extracts feature for each frame sampled from the video. Based on the features from two adjacent frames extracted by the feature generation sub-networks, a OFF sub-network is applied to generate the OFF for further classification. The scores from all sub-networks are fused to get the final result.}
\label{fig:overview_arch}
\end{figure}
% \vspace{-7px}

\label{sec:method}

Our proposed OFF is inspired by the famous brightness constant constraint defined by traditional optical flow~\cite{HornBertholdK.P.;Schunck1981}. It is formulated as follows:
\begin{equation}
\label{eq:constraint_image}
I(x, y, t) = I(x + \Delta x, y + \Delta y, t + \Delta t),
\end{equation}
where $I(x, y, t)$ denotes the pixel at the location $(x,y)$ of a frame at time $t$. For frames $t$ and $(t + \Delta t)$, $\Delta x$ and $\Delta y$ are the spatial pixel displacement in $x$ and $y$ axes respectively. It assumes that for any point that moves from $(x, y)$ at frame $t$ to $(x + \Delta x, y + \Delta y)$ at frame $t+\Delta t$, its brightness keeps unchanged over time.  
When we apply this constraint at the feature level, we have
\begin{equation}
\label{eq:constraint_feature}
% \begin{split}
f(I;w)(x, y, t) =f(I;w)(x + \Delta x, y + \Delta y, t + \Delta t),
% \end{split}
\end{equation}
where $f$ is a mapping function for extracting features from the image I. $w$ denotes the parameters in the mapping function. The mapping function $f$ can be any differentiable function. In this paper, we employ trainable CNNs consisted of stacks of convolution , ReLU, and pooling operations. According to the definition of optical flow, we assume that $p=(x, y,t)$ and obtain the equation as follows:
% The right-hand side of Equation~\ref{eq:constraint_feature} can be expanded by Taylor series as follows:
% \begin{equation}
% \label{eq:taylor}
% \begin{split}
% &f(I;w)(x + \Delta x, y + \Delta y, t + \Delta t) \approx f(I;w)(p) \\
%  &+\frac{\partial f(I;w)(p)}{\partial x}\Delta x + \frac{\partial f(I;w)(p)}{\partial y}\Delta y + \frac{\partial f(I;w)(p)}{\partial t}\Delta t,
%  \end{split}
% \end{equation}
% where $p=(x, y,t)$. From Equations ~\ref{eq:constraint_feature} and~\ref{eq:taylor}, we have
\begin{equation}
\label{eq:final_eq}
% \begin{split}
\frac{\partial f(I;w)(p)}{\partial x}\Delta x + \frac{\partial f(I;w)(p)}{\partial y}\Delta y + \frac{\partial f(I;w)(p)}{\partial t}\Delta t = 0.
%  \end{split}
\end{equation}
 By dividing $\Delta t$ in both sides of Equation~\ref{eq:final_eq}, we obtain
\begin{equation}
\label{eq:feature_flow_final}
% \begin{split}
\frac{\partial f(I;w)(p)}{\partial x}v_{x} + \frac{\partial f(I;w)(p)}{\partial y}v_{y} + \frac{\partial f(I;w)(p)}{\partial t} = 0,
%  \end{split}
\end{equation}
where $p=(x, y,t)$, and $(v_{x}, v_{y})$ denotes the two dimensional velocity of feature point at $p$. $\frac{\partial f(I;w)(p)}{\partial x}$ and $\frac{\partial f(I;w)(p)}{\partial y}$ are the spatial gradients of $\partial f(I;w)(p)$ in $x$ and $y$ axes respectively. $\frac{\partial f(I;w)}{\partial t}$ is the temporal gradient along time axis. 

As a special case, when $f(I;w)(p)=I(p)$, then $f(I;w)(p)$ simply represents pixel at $p$. In this special case, $(v_{x}, v_{y})$ are called optical flow. Optical flow is obtained by solving an optimization problem with the constraint in Equation~\ref{eq:feature_flow_final} for each $p$ \cite{barron1994performance,brox2004warpingflow,bigun1991multidimensional}. Here in this case, the term $\frac{\partial f(I;w)(p)}{\partial t}$ represents the difference between RGB frames.
Previous works have shown that the temporal difference between frames is useful in video related tasks \cite{wang2016tsn}, however, there is no theoretical evidence to help explain why this simple idea works that well. Here, we can find its correlation to spatial features and optical flow.

We generalize the representation of optical flow from pixel $I(p)$ to feature $ f(I;w)(p)$. In this general case, $[v_{x}, v_{y}]$ are called the feature flow.
We can see from Equation~\ref{eq:feature_flow_final} that $\vec{F}{\\}(I;w)(p)=[\frac{\partial f(I;w)(p)}{\partial x}, \frac{\partial f(I;w)(p)}{\partial y}, \frac{\partial f(I;w)(p)}{\partial t}]$ is orthogonal to the vector $[v_{x}, v_{y}, 1]$ containing feature-level optical flow. $\vec{F}{\\}(I;w)(p)$ changes as the feature-level optical flow changes. Therefore, $\vec{F}{\\}(I;w)(p)$ is guided by the feature-level optical flow. We call $\vec{F}{\\}(I;w)(p)$ as \textit{Optical Flow guided Feature (OFF)}. 
%When the condition $f(I;w)(p)=I(p)$ is satisfied, then we call $\vec{F}{\\}(I;w)(p)$ as the \textit{Raw OFF} in this case.
The OFF $\vec{F}{\\}(I;w)(p)$ encodes the spatial-temporal information orthogonally and complementarily to the feature-level optical flow $(v_{x}, v_{y})$.
In the next section, detailed implementation of OFF and its usage for action recognition are introduced. 

%For simplicity in implementation, we can use the Sobel operator on each channel of the feature map to generate the gradients $[\frac{\partial f(I;w)(p)}{\partial x}, \frac{\partial f(I;w)(p)}{\partial y}]$. And the $\frac{\partial f(I;w)(p)}{\partial t}$ could be implemented by an element-wise subtraction between the frames with time duration $\Delta t$.

\begin{figure*}[t]
\centering
\includegraphics[scale=0.9]{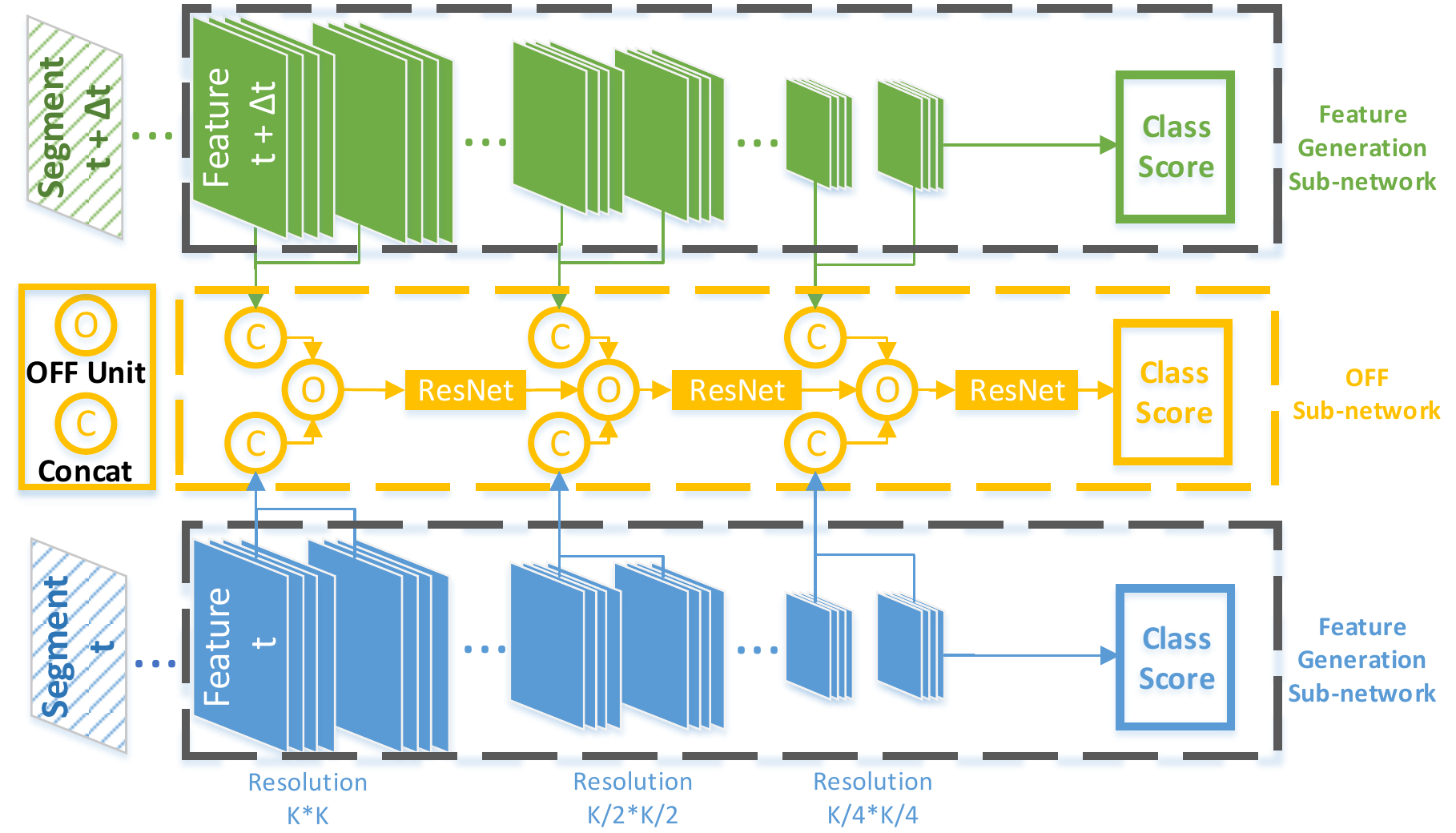}
% \bigskip
\caption{\textbf{Network architecture overview for two segments.} The inputs are two segments in blue and green colors that are separately fed into the feature generation sub-network to obtain basic features. In our experiment, the backbone for each feature generation sub-network is the BN-Inception \cite{Szegedycvpr2015googlenet}. Here K represents the largest side length of the square feature map selected to undergo the OFF sub-network for obtaining the OFF features. The OFF sub-network consists of several OFF units, and several residual blocks \cite{he2016resnet} are connected between OFF units from different levels of resolution. These residual blocks constitute a ResNet-20 when seen as a whole. The scores obtained by different sub-networks are supervised independently. Detailed structure of the OFF unit is shown in Figure \ref{fig:fag}.}
\label{fig:overview}
\end{figure*}

\section{Using Optical Flow Guided Feature in Convolutional Neural Network}
\label{sec:cnn}
%CNNs have been proved to be an excellent feature extractor for videos. The properties and advantages of CNNs, including invariance to illumination and noise, can be achieved by specific training strategy \cite{jaderberg2015stn}. In this paper, we utilize the CNNs as the mapping function $f(I)$ in the previous section. %Besides, we investigate the performance when applying the generating formulation defined in OFF among different input modalities, including RGB, RGB difference and optical flow respectively. We evaluate our method on two current state-of-the-art CNN frameworks, TSN and Two Stream Conv and. Both of them can achieve significant improvement in comparision with the baseline.

\subsection{Network Architecture}

 \textbf{Network Architecture Overview.} Figure \ref{fig:overview_arch} shows an overview of the whole network architecture. 
 The network consists of three sub-networks for different purposes: feature generation sub-network, OFF sub-network and classification sub-network. The feature generation sub-network generates basic features using common CNN structures. In the OFF sub-network, the OFF features are extracted using the features from the feature generation sub-network, and then several residual blocks are stacked for obtaining the refined features. The features from the previous two sub-networks are then used by the classification sub-network for obtaining the action recognition results. The Figure \ref{fig:overview} exhibits the more detailed network structure with the inputs of two segments. 
 As shown in Figure \ref{fig:overview}, we extract features from multiple layers on a specific level with the same resolution by concatenating them together and feed them into one OFF unit. The whole network has 3 OFF units with different scales. The details about the structure of each sub-network is discussed as follows.
 
 \vspace{-3px}
 \textbf{Feature Generation Sub-network.} %As we need to use the feature $f(I)$ extracted by CNN from modalities such as RGB, 
 The basic features $f(I)$ (equivalent to the representation $f(I;w)$ in previous section) are extracted from the input image using several convolutional layers with Rectified Linear Unit (ReLU) for non-linear function and max-pooling for down-sampling. We select BN-Inception  \cite{Szegedycvpr2015googlenet} as the network structure to extract feature maps. The feature generation sub-network can be replaced by any other network architecture.

\begin{figure}[t]
\centering
\includegraphics[scale=0.70]{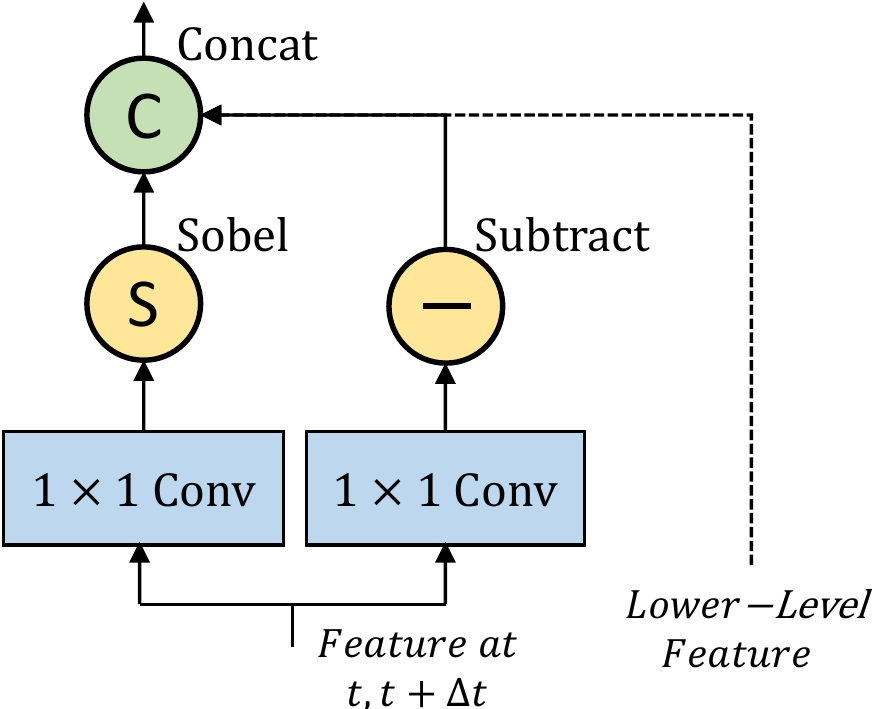}
\caption{\textbf{Detailed architecture of OFF unit.} A 1x1 convolution layer is connected to the input basic feature for dimension reduction. After that, we utilize the Sobel operator and element-wise subtraction to calculate the spatial and temporal gradients respectively. The combination of gradients constitutes the OFF, and the sobel operator, subtracting operator and the $1\times1$ convolution layers before them constitute a OFF layer.}
\label{fig:fag}
\end{figure}

  \textbf{OFF Sub-network.} The OFF sub-network consists of several OFF units. Different units use basic features $f(I)$ from different depths. As shown in Figure \ref{fig:fag}, an OFF unit contains an OFF layer to generate the OFF. Each OFF layer contains a $1\times1$ convolutional layer for each piece of feature, and a set of operators including sobel and element-wise subtraction for OFF generation. After the OFF is obtained, the OFF unit will concatenate them together with features from the lower level, then the combined features will be output to the following residual blocks.
 
%  \vspace{-6pt}
The OFF layer is responsible for generating the OFF from the basic features $f(I)$. Figure \ref{fig:fag} shows the detailed implementation the OFF layer. According to Equation~\ref{eq:final_eq}, the OFF should consist of both spatial and temporal gradient of the feature. Denote $f(I, c)$ as the $c$th channel of the basic feature $f(I)$. Denote $\mathcal{F}_x$ and $\mathcal{F}_y$ as the OFF for gradients of $x$ and $y$ directions respectively, which correspond to spatial gradients.
 We apply the Sobel operator for spatial gradient generation as follows:
 %Such operation feed the whole motion network with intermediate feature input, which may conduct vanishing gradient problem for the shallow layers of the network. 
%
% \begin{equation}
% % \begin{split}
% % &\frac{\partial f(I;w)(p)}{\partial x}\approx f_x(p) = Sobel_{x}(f(I;w)(p))\\
% % &=f(I;w)(x-1,y-1,t)-f(I;w)(x+1,y-1,t)\\
% % &\ \ +2f(I;w)(x-1,y,t)-2f(I;w)(x+1,y,t)\\
% % &\ \ +f(I;w)(x-1,y+1,t)-f(I;w)(x+1,y+1,t).
% % \end{split}
% % \[
% \begin{split}
% \mathcal{F}_{x} &= 
% \begin{Bmatrix}
% Sobel_{x} * f(I, c) | c = 0,1,...,N_{c}-1
% \end{Bmatrix}\\
% &=
% \begin{Bmatrix}
% \begin{bmatrix}
% -1 & 0 & 1 \\
% -2 & 0 & 2 \\
% -1 & 0 & 1 
% \end{bmatrix} * f(I, c) | c = 0,1,...,N_{c}-1
% \end{Bmatrix}
% \end{split}
% \end{equation}
%
\begin{equation}
\mathcal{F}_{x} = \left\{ \begin{bmatrix}
-1 & 0 & 1 \\
-1 & 0 & 1 \\
-1 & 0 & 1 
\end{bmatrix} * f(I, c) \bigg\vert  c = 0 \ldots, N_c - 1 \right\} 
\end{equation}
\begin{equation}
\mathcal{F}_{y} = \left\{ \begin{bmatrix}
1 & 1 & 1 \\
0 & 0 & 0 \\
-1 & -1 & -1
\end{bmatrix} * f(I, c)　\bigg\vert c = 0, \ldots, N_c - 1 \right\}
\end{equation}
%
% \begin{align}
% \mathcal{F}_{x} &= 
% \begin{Bmatrix}
% Sobel_{x} * f(I, c) | c = 0,1,...,N_{c}-1
% \end{Bmatrix}\\
% &=
% \left\lbrace
% \begin{bmatrix}
% -1 & 0 & 1 \\
% -2 & 0 & 2 \\
% -1 & 0 & 1 
% \end{bmatrix} * f(I, c) | c = 0,1,...,N_{c}-1 \right\rbrace
% \end{align} 
% 
% \begin{equation}
% % \begin{split}
% % &\frac{\partial f(I;w)(p)}{\partial y}\approx f_y(p) = Sobel_{y}(f(I;w)(p))\\
% % &=f(I;w)(x-1,y-1,t)-f(I;w)(x-1,y+1,t)\\
% % &\ \ +2f(I;w)(x,y-1,t)-2f(I;w)(x,y+1,t)\\
% % &\ \ +f(I;w)(x+1,y-1,t)-f(I;w)(x+1,y+1,t).
% % \end{split}
% \begin{split}
% \mathcal{F}_{y} &= \begin{Bmatrix}
% Sobel_{y} * f(I, c) | c = 0,1,...,N_{c}-1
% \end{Bmatrix}\\
% &= 
% \begin{Bmatrix}
% \begin{bmatrix}
% 1 & 2 & 1 \\
% 0 & 0 & 0 \\
% -1 & -2 & -1
% \end{bmatrix} * f(I, c) | c = 0,1,...,N_{c}-1
% \end{Bmatrix}
% \end{split}
% \end{equation}
 where $\ast$ denotes a convolution operation, and the constant $N_{c}$ indicates the number of channels of the feature $f(I)$. 
 Denote $\mathcal{F}_t$ as the OFF for gradients at the temporal directions.  Temporal gradient is obtained by element-wise subtraction as follows:
%  \begin{equation}
% \mathcal{F}_{t} = \{(f(I, c)(x,y,t) -\\ f(I, c)(x,y,t-\Delta t)) | c = 0,1,...,N_c-1\}
% \end{equation}
%
\begin{equation}
\mathcal{F}_{t} = \{ f_{t}(I, c) - f_{t-\Delta t}(I, c) | c = 0, \ldots, N_c-1\}
\end{equation}

With the features $\mathcal{F}_x$, $\mathcal{F}_y$, and $\mathcal{F}_t$ obtained above, we concatenate them together with the features from the lower level as the output of the OFF layer. We use a $1\times 1$ convolutional layer before the sobel and subtraction operations to reduce the number of channels. In our experiments, the channel dimension is reduced to 128 regardless of how many the input channels are. Then the feature is fed into the OFF unit to calculate the OFF we defined in previous section.
%Given input of size $W\times H \times C$ defined by the $1\times1$ convolutional layer for dimensionality reduction, the output of the OFF layer is $W\times H \times 3C$. Therefore, the spatial resolution keeps unchanged while the number of channels for the output is 3 times the number for the input.
 After the OFF is obtained, several residual blocks designed in \cite{he2016resnet} are connected between the OFF units at different levels of resolution as refinement. The dimensionality of OFF is further reduced in the residual block adjacent to the OFF unit for saving computation and the number of parameters. The residual blocks on different levels of resolution finally constitute a ResNet-20. Note that there is no Batch Normalization \cite{ioffe2015batchnorm} operation applied in our residual network in order to avoid the over-fitting problem.

The OFF unit can be applied for CNN layers on different levels.
 The inputs of one OFF unit include the basic deep features from two segments, and the feature from the OFF unit on the previous feature level if it exists. In this way, the OFF at the previous semantic level can be used for refining the OFF at the current semantic level.

%\begin{equation}
%\begin{split}
%\mathcal{F}_{l}(f(I)_1, f(I)_2, &..., f(I)_K)=\\&\{Sobel(f(I)_1, %f(I)_2, ..., f(I)_K), \\&T(f(I)_1, f(I)_2, ..., f(I)_K)\}
% \end{split}
%\end{equation}

\textbf{Classification Sub-network.}
The classification sub-network takes features from different sources and uses multiple inner-product classifiers to obtain multiple classification scores. The classification scores of all sampled frames are then combine by averaging for each feature generation sub-network, or OFF sub-network. The OFF at a semantic level can be used to produce a classification score at the training stage, which is learned using its corresponding loss. Such strategy has been proved to be useful in many tasks \cite{Szegedycvpr2015googlenet,wei2016cpm,newell2016hourglass}. In the testing phase, scores from different sub-networks could be assembled for better performance.

% and denote the $\tilde{y}_l$ as the classification function for $\mathbf{G}_l$, the classification procedure could be formulated as:

% \begin{equation}
% \begin{split}
% \tilde{y}_l = \mathbf{W}_{c,l}\mathcal{G}_{l},
%  \end{split}
% \end{equation}
% where $\mathbf{W}_{c,l}$ is the learned classifier implemented by a fully connected layer.

\vspace{-8px}
\subsection{Network Training}
Action recognition is treated as a multi-class classification problem. 
% {\color{red}
Followed by the settings in TSN, as there are multiple classification scores produced by each segment, we need to fuse them all in each sub-network separately to generate a video-level score for loss calculation. Here, for the OFF sub-networks, the features produced by the output of OFF sub-network for the $t$th segment on level $l$ is denoted by $\mathcal{F}_{t, l}$. The classification score for segment $t$ on the level $l$ using $\mathcal{F}_{t, l}$ is denoted by $\mathbf{G}_{t, l}$. The aggregated video-level score at level $l$ is denoted by $G_{l}$. The video-level action classification score $G_{l}$ is obtained by:
%Similarly, we can also use the features from the last layer of the feature generation sub-network to obtain classification scores, which is denoted by $\tilde{y}_t$. The final action classification score $\tilde{y}_{final}$ is obtained by:
 \begin{equation}
% \begin{split}
G_{l} = \mathcal{G}(\mathbf{G}_{0, l}, \ldots, \mathbf{G}_{1, l}, \ldots, \mathbf{G}_{N_{t-1}-1, l}),
%  \end{split}
\end{equation}
where $N_t$ denotes the number of frames for extracting features. The aggregation function denoted by $\mathcal{G}$ is used for summarizing the scores predicted from different segments along time. Following the investigations in TSN, $\mathcal{G}$ is implemented by average pooling for better performance \cite{wang2016tsn}. As for the feature generation sub-network, the above equations are also applicable. While as we do not need intermediate supervision for feature generation sub-network, the feature $\mathcal{F}_{t, l}$ at level $l$ for segment $t$ is simply equivalent to the final feature output of the sub-network.

To update the parameters of the whole network, the loss is set to be the standard categorical cross-entropy loss. As the sub-network for each feature level is supervised independently, a loss function is used for each level as:
\begin{equation}
% \begin{split}
 \mathcal{L}_{l}(y, G_{l})=-\sum_{c=1}^{C}y_{c}(G_{l, c} - \log\sum_{j=1}^{C}e^{G_{l, j}}).
%Wanli's: {L}=-\sum_{l=1}^L \sum_{i=1}^{C}y_{t,i} \log{\tilde{y}_{t,i}^l})
%  \end{split}
\end{equation}
where $C$ is the number of action categories, $G_{l, c}$ is the estimated score for class $c$ from the features at level $l$, and $y_c$ represents the ground-truth class label.
% }
By using this loss function we can optimize the network parameters through back-propagation. Detailed implementation of training is described as follows.

% \vspace{-3px}
\textbf{Two-stage Training Strategy.} %Mentioned before, deleted here. The FEN sub-network could be constructed on any kinds of CNN architectures, and we use the BN-Inception as the back-bone in this paper. 
Training of the whole network consists of two stages. The first stage indeed is to apply existing approaches, e.g. TSN \cite{wang2016tsn}, to train the feature generation sub-network. %For TSN, we use the same strategy as \cite{wang2016tsn}, including sampling methods, pre-processing and data augmentation. 
At the second stage, we train the OFF and classification sub-network with all the weights in feature generation sub-network frozen. The weights of OFF sub-network and classification sub-network are learned from scratch. 
The whole network could be further fine-tuned in an end-to-end manner, however, we do not find significant gain in this stage. To simplify the training process, we only train the network using the first two stages.
% Other parameters that are not fixed in the original framework will be introduced in the following experimental section.

% \vspace{-3px}
\textbf{Intermediate Supervision during Training.} Intermediate supervision has been proven to be practical training strategy in many other computer vision tasks \cite{newell2016hourglass, wei2016cpm, Yangwei_2017_ICCV, Ouyang_2017_ICCV,Chu_2017_CVPR}. As the OFF sub-networks are fed by intermediate inputs, here we add the intermediate supervision on each level to get better OFFs on each level of resolution.

% However, the scores obtained in shallow levels would not serve as the classification results in testing.

%Note that we follow the practices in very deep Two-Stream ConvNets \cite{wang2015verydeep}, which is the state-of-the-art implementation of the two-stream framework.

%\textbf{Data Augmentation.} 
%(basic knowledge, no need to mention here) Data augmentation is a technique which generates huge amount of training samples from a limited training dataset. Usually when there comes overfitting problems, heavy data augmentation is required to expand the fitness of the model. 
%We us the same data augmentation approach, i.e. flipping, corner cropping and scale jittering, as that in the TSN \cite{wang2016tsn}. We adopt all these useful methods to train and compare with TSN under the same testing circumstance. 
%While testing under Two-Stream ConvNet architecture, those techniques introduced in TSN will also be applied in order to keep the irrelevant parameters invariant between ours and baseline.

% \textbf{Train-Test Interval Consistency.} 
% \vspace{-3px}
\textbf{Reducing the Memory Cost.} 
As our framework consists of several sub-networks, it costs more memory than the original TSN framework, which extracts and stores motion frames before training CNNs, and trains several networks independently. In order to reduce the computational and memorial cost, we sample less frames in the training phase than in the testing phase, and still obtain satisfactory results.

However, the time duration between segments may be varied if we sample different number of segments between training and testing. According to our definition in equation \ref{eq:final_eq}, only when the denotation $\Delta t$ is a fixed constant, the equation \ref{eq:feature_flow_final} could be derived from the equation \ref{eq:final_eq}. If we sample different frames between training and testing, the time interval $\Delta t$ may be inconsistent, which makes our definition to be invalid and influences the final performance. In order to keep time interval consistent between training and testing, we design the sampling scheme carefully. Therefore, during training, we sample frames from a video as follows:

% As we only sample 3 frames as training samples during training, while in testing we sample 25 frames per video. In order to keep the frame temporal interval consistent between training and testing, we sample the frames during training as follows.
Let $\alpha$ be the number of frames sampled for training, and $\beta$ be the number for testing. In training phase, a video with length $L, L>=\beta$ would be divided into $\beta$ segments. Each segment has length $\floor*{L / \beta}$. We randomly select $p$ from ${0, 1, \ldots, L - 1 - (\alpha-1) * \floor*{L / \beta}}$, where $p$ is treated as a frame seed. Then the whole training set is constructed as $\{p, p + \floor*{L / \beta}, ..., p + (\alpha - 1) * \floor*{L / \beta}\}$, which has interval $\floor*{L / \beta}$. In testing phase,  we sample the images using the same interval $\floor*{L / \beta}$ as that in the training phase.

\subsection{Network Testing}

% As there are multiple classification scores produced by different sub-networks for each segment, we need to fuse them all in testing phase for better performance. Here, the score produced by the output of OFF sub-network for the $t$th segment is denoted by $\mathcal{F}_{t}$. The classification score for segment $t$ using $\mathcal{F}_{t}$ is denoted by $\mathbf{G}_{t}$. Similarly, we can also use the features from the last layer of the feature generation sub-network to obtain classification scores, which is denoted by $\tilde{y}_t$. The final action classification score $\tilde{y}_{final}$ is obtained by:
%  \begin{equation}
% % \begin{split}
% \tilde{y}_{final} = \mathcal{G}(\tilde{y}_0, \ldots, \tilde{y}_{N_t-1}, \mathbf{G}_{0}, \ldots, \mathbf{G}_{1}, \ldots, \mathbf{G}_{N_{t-1}-1}),
% %  \end{split}
% \end{equation}
% where $N_t$ denotes the number of frames for extracting features. The aggregation function denoted by $\mathcal{G}$ is used for summarizing the scores predicted from different frames. Following the investigations in TSN, $\mathcal{G}$ is implemented simply by average pooling \cite{wang2016tsn}. The fused score will be served as the final output of the whole set of networks.

% {\color{red}
As there are multiple classification scores produced by different sub-networks, we need to fuse them all in testing phase for better performance. In this study, we  assemble scores from the feature generation sub-network and the last level of OFF sub-network by a simple summing operation. We select to test our model based on a state-of-the-art framework TSN \cite{wang2016tsn}. The testing setting under the TSN framework is illustrated as follows:
% }

% \textbf{Testing for the Two-Stream Framework.} In Two-Stream Framework \cite{simonyan2014two}, 25 frames are sampled equidistantly from each video. In order to capture spatial temporal information from such large displacement, each OFF feature is directly generated from a consecutive frame pair among those sampled images. In this case, we sample 25 independent frames and generate 24 OFF maps using the OFF sub-network at each level. Other settings are kept the same as that in the two-stream ConvNet in order to fairly compare with the Two-Stream approach in the experiment result. 

\textbf{Testing under TSN Framework.} 
%The only difference in testing process between TSN \cite{wang2016tsn} and Two-Stream ConvNet \cite{simonyan2014two} are the sampling approach and final segment consensus methodology.
In the testing stage of TSN, 25 segments are sampled from RGB, RGB difference, and optical flow. However, the number of frames in each segment is different among these modalities. We use the original settings adopted by TSN  to sample 1, 5, 5 frames per segment for RGB, RGB difference, and optical flow respectively. The input of our network is 25 segments, where the $t$th segment is treated as the Frame $t$ in Figure \ref{fig:overview}. In this case, the features extracted by a separate branch of our feature generation sub-network is for a segment instead of a frame when using TSN. Other settings are kept to be the same as those in TSN.

\vspace{-5px}
\section{Experiments and Evaluations}
\label{sec:experiment}
In this section, datasets and implementation details used in experiments will be first introduced. Then we will explore the OFF and compare it with other modalities under current state-of-the-art frameworks. Moreover, as our method can be extended to other modalities such as RGB difference and optical flow, we will show how such a simple operation could improve the performance for input with different modalities. Finally, we will discuss the meaning and difference between the OFF and other motion modalities such as optical flow and RGB difference.

\subsection{Datasets and Implementation Details}
\textbf{Evaluation Datasets.} The experimental results are evaluated on two popular video action datasets, UCF-101 \cite{khu2012ucf101} and HMDB-51 \cite{Kuehne11hmdb}. The UCF-101 dataset has 13320 videos and is divided into 101 classes, while the HMDB-51 contains 6766 videos and 51 classes. Our experiments follow the officially offered scheme which divides a dataset into 3 training and testing splits and finally calculating the average accuracy over all 3 splits. We prepare the optical flow between frames before training by directly using the OpenCV implemented algorithm \cite{zach2007tvl1}.

\textbf{Implementation Details.} We train our model with 4 NVIDIA TITAN X GPU, under the implementation on Caffe \cite{jia2014caffe} and OpenMPI. We first train the feature generation sub-networks using the same strategy provided in the corresponding method \cite{wang2016tsn}. 
Then at the second stage, we train the OFF sub-networks from scratch with all parameters in the feature generation sub-networks frozen. The mini-batch stochastic gradient descent algorithm is adopted here to learn the network parameters. When the feature generation sub-networks are fed by RGB frames, the whole training procedure for OFF sub-network takes 20000 iterations to converge with the learning rate initialized at 0.02 and decreased to its 0.1 using multi-step policy at the iteration 10000, 15000 and 18000. When input changes to temporal modality like optical flow, the learning rate is initialized at 0.05, and other policies are kept the same with what have been proposed in RGB. The batch size is set to 128 and all the training strategies described in previous sections are applied. When evaluating on UCF-101 and HMDB-51, we add dropout modules on spatial stream of OFF.
%Finally at the third stage, we fine-tune the whole network end-to-end with the learning rate of 0.0002 and 0.0005 respectively for RGB and optical flow. %for 2000 iterations.
%When training for the TSN framework, we fix the segment number as 7 in both spatial and temporal networks, and the segment length of temporal network is 5, which means that each segment contains 5 continuous frames.% As for network structure settings, we set the number of channels of the $1\times1$ conv layer to be 128,.
There is no difference on training parameters for different modalities. However, when the input is RGB difference or optical flow, it would cost more time in both training and testing stages as more frames are read into the network.

\subsection{Experimental Investigations on OFF.}

In this section, we will investigate the performance of OFF under the TSN framework. The analysis for the performance of single and multiple modalities, and the performance comparison between the state-of-the-art will be shown. All the results for OFF based networks are trained with the same network backbone and strategies illustrated in previous sections for fair comparison.

% \begin{center}
\begin{table}

\setlength{\tabcolsep}{8.8pt}
\begin{tabular}{lcc}
 \Xhline{3\arrayrulewidth}
Method & Speed (fps) &
Acc. \\
\hline
 
 \hline
 TSN(RGB) \cite{wang2016tsn} & 680 & 85.5\% \\
 \hline
 TSN(RGB+RGB Diff) \cite{wang2016tsn} & 340 & 91.0\% \\
 \hline
 TSN(Flow) \cite{wang2016tsn} & 14 & 87.9\%  \\
 \hline
 TSN(RGB+Flow) \cite{wang2016tsn} & 14 & 94.0\% \\
 \hline
 \makecell[l]{RGB+EMV-CNN \cite{zhang2016motionvector}} & 390 & 86.4\% \\
 \hline
 \makecell[l]{MDI+RGB \cite{bilen2016dynamic}} & \textless131 & 76.9\% \\
 \hline
%  \makecell[l]{Hidden Two-Stream \cite{zhu2017hidden}} & 120 & 89.8\% \\
%  \hline
 \makecell[l]{Two-Stream I3D\\(RGB+Flow) \cite{carreira2017i3d}} & \textless14 & 93.4\% \\
 \Xhline{3\arrayrulewidth}
%  RGB+Raw OFF  & 340 & 90.7\%\\
%  \hline
%  RGB+OFF(RGB)& 450 & 90.0\% \\
%  \hline
 \makecell[l]{RGB+OFF(RGB)+\\RGB Diff+OFF(RGB Diff)}& 206 & \textbf{93.3\%} \\
 \Xhline{3\arrayrulewidth}

\end{tabular}
% \bigskip
% \vspace{3pt}
\caption{Experimental results of accuracy and efficiency for different real-time video action recognition methods on \textit{UCF-101 over three splits}. Here the notation \textit{Flow} represents the motion modality Optical Flow. Note that our OFF based algorithm could achieve the state-of-the-art performance among real-time algorithms.}
\label{table:efficiency}
 \end{table}
% \end{center}

% \setlength{\parskip}{1pt}
\textbf{Efficiency Evaluation.} In this experiment, we evaluate the efficiency between the OFF based method and other state-of-the-art methods. The experimental results for efficiency and accuracy for different algorithms are summarized in Table \ref{table:efficiency}. OFF(RGB) denotes our use of OFF for the network with RGB input, in this case, the OFF is acquired from spatial deep features. As one special case, the denotation \textit{RGB Diff} represents the OFF calculated directly from consecutive RGB frames on the input level instead of on the feature level. After applying the OFF calculation to RGB frames, the processed inputs could be fed into the feature generation sub-network and the generated feature maps could be again used to calculate their corresponding OFF features on the feature level.
The other methods we compared here includes TSN \cite{wang2016tsn} with different inputs, motion vector based RGB+EMV-CNN \cite{zhang2016motionvector}, dynamic image based CNN \cite{bilen2016dynamic} and current state-of-the-art 3D-CNN with two stream \cite{carreira2017i3d}.
From the Table \ref{table:efficiency}, by applying the OFF to the spatial features and the RGB inputs, we can achieve a competitive accuracy $93.3\%$ with only RGB inputs on the UCF-101 over three splits, which is even comparable with some Two-Stream based methods such as \cite{carreira2017i3d,wang2016tsn}. Besides, our methods is still very efficient under this kind of settings. The whole network could run over 200 fps, while other methods listed here are either inefficient or not so effective as the Two-Stream based approaches.
%When single modality is used, the result exhibits that OFF has the best performance compared with the raw OFF, optical flow and motion vector, while it can run on-the-fly at 450 fps, which is 32 times the speed of optical flow. Moreover, when fusing the spatial CNNs and temporal CNNs, our OFF add very small extra cost, because feature generation sub-networks share most weights with the spatial CNNs. However, other motion-based representations like optical flow and raw OFF have double computational cost, as the input of the temporal CNNs are totally different.
% Moreover, when fusing those results with RGB scores, the motion-based modalities like optical flow and RGB-difference need to set up another independent network to train with the RGB to serve as the input.
%\textbf{RGB+Motion Evaluation.\textcolor{red}{consider remove this section}}
%As introduced in \cite{simonyan2014two}, fusing the result of RGB and motion information would obtain higher accuracy. Therefore, we sum the score from the RGB network with the score from different implementations of obtaining temporal information. As shown by the results in the second column of Table \ref{table:single}, 

% \begin{center}
\begin{table}

\setlength{\tabcolsep}{1.8pt}
\begin{tabular}{|cccccccc|}
\hline
RGB & \makecell{OFF\\(RGB)} & \makecell{RGB\\Diff}  & \makecell{OFF\\(RGB Diff)} & Flow & \makecell{OFF\\(Flow)} &
\makecell{Speed \\(fps)} & \makecell{Acc.} \\
 \hline
 \checkmark & & & & & & 680 & 85.5\% \\
 %\hline
 \checkmark & \checkmark & & & & & 450 & 90.0\% \\
 \hline
 \checkmark & & \checkmark & & & & 340 & 90.7\% \\
 %\hline
 \checkmark & \checkmark & \checkmark & & & & 257 & 92.0\% \\ % 92.38
 %\hline
 \checkmark & \checkmark & \checkmark & \checkmark & & & 206 & 93.0\% \\
 \hline
 \checkmark & & & & \checkmark & & 14 & 93.5\% \\
 %\hline
  \checkmark & \checkmark & & & \checkmark &  & 14 & 95.1\% \\ % 94.69
 %\hline
 \checkmark & \checkmark & & & \checkmark & \checkmark & 14 & 95.5\% \\ %94.97
 \hline
\end{tabular}
% \bigskip
% \vspace{1pt}
\caption{Experimental results for different modalities using the OFF on \textit{UCF-101 Split1}. Here Flow denotes the optical flow. OFF(*) denotes the use of OFF for the input *. For example, OFF(RGB) denotes the use of OFF for RGB input. The speed here illustrates the time cost for network forward. The results for RGB and RGB + Flow are from \cite{wang2016tsn}. The OFF(RGB) provides a strong $4.5\%$ improvement when fusing with RGB.}
\label{table:compleval}
\end{table}
% \end{center}

\textbf{Effectiveness Evaluation.} In this part, we try to investigate the robustness of OFF when applying to different kinds of input. According to the definition in equation \ref{eq:feature_flow_final}, we can replace the image $I$ from RGB image to optical flow or RGB difference image to extract OFF on feature level for further experiments. Based on the scores predicted by different modalities, we can further improve the classification performance by fusing them together \cite{simonyan2014two, diba20163dtwostream, wang2016tsn, zhang2016motionvector}. We carry out the experimental results with various score fusing schemes on UCF-101 split 1, and summarize them in Table \ref{table:compleval}. Table \ref{table:compleval} shows the results when different kinds of modalities are introduced as the network input. From each block separated by a horizon line, we can find that the OFF is complementary to other kinds of modalities, e.g. RGB and optical flow, and could get a remarkable gain every time the OFF is introduced. Besides, interestingly, the OFF is still working when the input modality is already describing the motion information. This phenomenon indicates that the \textit{acceleration information} between frames might also make a difference in describing the temporal patterns. 
%These experiments show that OFF is complementary to other kinds of spatial and temporal information like RGB and optical flow.

%\textcolor{red}{By using the OFF both on feature and input level, we  achieve the accuracy at $93.0\%$ on UCF101 split1, which is comparable to the current state-of-the-art spatial-temporal 3D based approach \cite{carreira2017i3d} and TSN with optical flow \cite{wang2016tsn}.} %While the model is running on-the-fly with 220 frames per second, the performance can even outperform many optical-flow based methods, which could only run at 14fps.

% \begin{center}
\begin{table}

\setlength{\tabcolsep}{5.8pt}
\begin{tabular}{cccc}
\Xhline{2\arrayrulewidth}
  & RGB & \makecell{Hyp-Net + RGB} & \makecell{OFF(RGB) + RGB} \\
\hline
 
 Acc. & 85.5\% & 86.0\% & 90.0\% \\
\Xhline{2\arrayrulewidth}

\end{tabular}
% \bigskip
% \vspace{5pt}
\caption{Experimental results of accuracy for hypercolumn network and the comparison with OFF on UCF-101 Split1. The denotation "Hyp-Net" indicates the output of hypercolumn network.}
\label{table:hyper}
 \end{table}
% \end{center}

\textbf{Comparison with the Hypercolumns CNN}. As our network extracts intermediate deep features from a pre-trained CNN, such \textit{hypercolumn} based network structure may lead to additional gain on specific datasets \cite{hariharan2015hypercolumns}. Experiment and analysis are conducted  to investigate whether the OFF is playing a key role for the improvement. The network architecture and all training strategies for the hypercolumn CNN are the same as that in OFF except for the removal of OFF unit, in other words, the hypercolumn network here is constructed as the same structure of OFF sub-network without OFF unit. In this case, the features from feature generation sub-networks are directly fed into the OFF sub-networks without the calculation of OFF.

From the experimental results shown in Table \ref{table:hyper}, it is clear that, despite the hypercolumn network could get a slight $0.5\%$ improvement on UCF-101 split 1, its final accuracy is still apparently less than the one obtained by OFF(RGB). Therefore, a conclusion could be drawn that it is the OFF calculation rather than the hypercolumn  structure that plays the key role in achieving the significant gain.

\textbf{Comparison with the State-of-the-art.} Above all, after the exploration and analysis of the OFF, we show our final result. As what has been done in TSN, we also assemble the classification scores obtained by different kinds of modalities. We sum the scores produced by each modality together, and get the final version output in Table \ref{table:final}. All the results are evaluated in the UCF-101 and HMDB-51 over 3 splits. Our results are obtained by assembling the scores from RGB, OFF(RGB), optical flow and their corresponding version of OFF(optical flow) together. When we add one more score from OFF(RGB Diff), a slight 0.3\% gain is obtained compared to the version without it, and finally results in $96.0\%$ on UCF-101 and $74.2\%$ on HMDB-51. Note that we do not introduce improved Dense Trajectories (iDT)\cite{wang2013idt} into our network as the input. The components of inputs we need to prepare in advance  for our final version result only consist of RGB and optical flow.
%, which could be regarded as a two-stream approach. 

We compare our result with both the traditional approaches and deep learning based approaches. We obtain $2.0\%/5.7\%$ gain compared with the baseline Two-Stream TSN \cite{wang2016tsn} on UCF-101 \cite{khu2012ucf101} and HMDB-51 \cite{Kuehne11hmdb} respectively. Note that the final version TSN takes 3 modalities (RGB, Optical Flow and iDT) as network input. The other compared methods listed in Table \ref{table:final} include iDT \cite{wang2013idt}, Two-Stream ConvNet \cite{simonyan2014two}, Two-Stream + LSTM \cite{yue2015beyondshortlstm}, Temporal Deep-convolutional Descriptors (TDD) \cite{wang2015tdd}, Long-term Temporal Convolutions (LTC) \cite{varol2017ltc}, Key Volume Mining Deep Framework (KVMDF) \cite{zhu2016kvmf}, and also the current state-of-the-art methods such as Spatio-Temporal Pyramid (STP) \cite{wang2017spatiotemporalpyramid}, Saptio-Temporal Multiplier Network (STMN) \cite{feichtenhofer2017spatio-temp-multiplier}, Spatio-Temporal Vector \cite{cosmin2017stvector}, Lattice LSTM (L$^2$STM) \cite{sun2017lattice}, and I3D \cite{carreira2017i3d}. The method I3D could achieve spectacular performance (98.0\% on UCF-101, 80.7\% on HMDB-51, over 3 splits) when proposing a new large dataset \textit{Kinetics} for pre-train. While without the pre-training, the method I3D could achieve 93.4\% on UCF-101 Split1. From the comparison with all the listed methods, we conclude that our OFF based method allow for state-of-the-art performance in video action recognition.

% \begin{center}
\begin{table}[t]

\setlength{\tabcolsep}{8pt}
\begin{tabular}{|l||c|c|c|c|}
\hline
\makecell{Method} &
UCF-101 & \makecell{HMDB-51} \\
\hline
 iDT \cite{wang2013idt} & 86.4\% & 61.7\% \\
 \hline
 Two-Stream \cite{simonyan2014two} & 88.0\% & 59.4\% \\
 \hline
 Two-Stream TSN \cite{wang2016tsn}  & 94.0\% & 68.5\% \\
 \hline
 Three-Stream TSN \cite{wang2016tsn}  & 94.2\% & 69.4\% \\
 \hline
 Two-Stream+LSTM \cite{yue2015beyondshortlstm} & 88.6\% & -\% \\
 \hline
 TDD+iDT \cite{wang2015tdd}  & 91.5\% & 65.9\% \\
 \hline
 LTC+iDT \cite{varol2017ltc}  & 91.7\% & 64.8\% \\
 \hline
 KVMDF \cite{zhu2016kvmf}  & 93.1\% & 63.3\% \\
 \hline
 STP \cite{wang2017spatiotemporalpyramid}  & 94.6\% & 68.9\% \\
 \hline
 STMN+iDT \cite{feichtenhofer2017spatio-temp-multiplier}  & 94.9\% & 72.2\% \\
 \hline
 ST-VLMPF+iDT \cite{cosmin2017stvector} & 94.3\% & 73.1\% \\
 \hline
 L$^2$STM \cite{sun2017lattice} & 93.6\% & 66.2\% \\
 \hline
 Two-Stream I3D \cite{carreira2017i3d}  & 93.4\% & 66.4\% \\
 \hline
 \makecell[l]{Two-Stream I3D \\(with Kinetics 300k) \cite{carreira2017i3d}}  & 98.0\% & 80.7\% \\
 \hline
% \makecell{I3D (with extra data)}  & 98.0\% & 80.7\% \\
% \hline
% \makecell{I3D (without extra data)}  & 93.4\% & ??\% \\
% \hline
 %\makecell{I3D (with Kinetics pretrain)}  & 98.0\% & ~\% \\
 %\hline
 %\makecell{TLE (on 3DCNN)}  & 95.0\% & ~\% \\
 %\hline
 %\makecell{TLE (on 2DCNN)}  & 95.0\% & ~\% \\
 %\hline
 \makecell{\textbf{Ours}} & \textbf{96.0\%} & \textbf{74.2\%}\\ % 95.
 \hline
%   \makecell{ours: RGB+OFF (RGB) +\\OF+OFF (OF)+OFF(RGB-Diff)} & 95.3\% & -\%\\
%  \hline

\end{tabular}
% \bigskip
% \vspace{5pt}
\caption{Performance comparison to the state-of-the-art methods on UCF-101 and HMDB-51 over 3 splits.}
\label{table:final}
 \end{table}
% \end{center}

\vspace{-5px}
\section{Conclusion}

\label{sec:con}

In this paper, we have presented \textit{Optical Flow guided Feature (OFF)}, a novel motion representation derived from and guided by the optical flow. OFF is both fast and robust. By plugging the OFF into CNN framework, the result with only RGB as input on UCF-101 is even comparable to the result obtained by Two-Stream (RGB+Optical Flow) approaches, and at the same time, the OFF plugged network is still very efficient with the speed over 200 frames per second. Besides, it has been proven that the OFF is still complementary to other motion representations like optical flow. Based on this representation, we proposed an new CNN architecture for video action recognition. This architecture outperforms many other state-of-the-art video action recognition methods on two popular video datasets UCF-101 and HMDB-51, and could be used to accelerate the speed of the video based tasks. In future works, we will validate our method on other video based tasks and datasets.
%As a general motion representation, the OFF has potential expandability on other video based tasks. We will investigate this method on other tasks and datasets in the future.
%In this paper, we presented a new spatial temporal representation - Orthogonal Feature Flow (OFF). The representation is derived based on the definition of optical flow, and is fast and easy to obtain. Besides, while running in an extremely fast speed, the modality could also achieve comparable accuracy even when compared with the dense motion modality like optical flow. Previous work has provided precious knowledge about network structure, sampling methods and training strategies, and in this work we utilized all of these experience and applied our method on different frameworks to achieve performance improvement. This also proves that our method can be applied in different video action recognition frameworks. We also present the acceleration-based information based on the design philosophy of OFF in video action recognition tasks, which achieved the state-of-the-art performance on two challenging datasets respectively. In the future work, we will explore using our method for other video based tasks.

%\section{Acknowledgement}

{\small
\bibliographystyle{ieee}
\bibliography{references}
}
\end{document}